\documentclass[letterpaper]{article}

\usepackage{natbib,alifeconf}
\usepackage{url,hyperref,cleveref}
\usepackage{booktabs}

\usepackage{standalone}
\usepackage{tikz}
\usetikzlibrary{arrows, backgrounds, calc, fit, positioning, shapes.geometric}

\usepackage{subcaption}

\title{Interactive embodied evolution for socially adept Artificial General Creatures}

\author{
    Kevin Godin-Dubois$^{1}$,
    Oliver Weissl$^{1}$,
    Karine Miras$^{1}$, \and
    Anna V. Kononova$^2$ \\
    \mbox{}\\
    $^1$Vrije Universiteit Amstderdam, Netherlands \\
    $^2$Leiden University, Netherlands \\
    k.j.m.godin-dubois@vu.nl
}

\begin{document}

\maketitle

\begin{abstract}
 We introduce here the concept of Artificial General Creatures (AGC) which encompasses "robotic or virtual agents with a wide enough range of capabilities to ensure their continued survival".
 With this in mind, we propose a research line aimed at incrementally building both the technology and the trustworthiness of AGC.
 The core element in this approach is that trust can only be built over time, through demonstrably mutually beneficial interactions.
 
 To this end, we advocate starting from unobtrusive, nonthreatening artificial agents that would explicitly collaborate with humans, similarly to what domestic animals do.
 By combining multiple research fields, from Evolutionary Robotics to Neuroscience, from Ethics to Human-Machine Interaction, we aim at creating embodied, self-sustaining Artificial General Creatures that would form social and emotional connections with humans.
 Although they would not be able to play competitive online games or generate poems, we argue that creatures akin to artificial pets would be invaluable stepping stones toward symbiotic Artificial General Intelligence.
 \end{abstract}

As far back as the mid-1950s, researchers were foreseeing a future where Artificial General Intelligence (AGI) would be commonplace.
For example, in \citep{Simon1958} the authors state that ``within ten years a digital computer will write music [with] considerable aesthetic value".
However, more than half a century later, AGI is still a much-investigated \emph{theoretical} problem \citep{Pennachin2007, Goertzel2014, Roli2022}.
While some of the recent advances in Large Language Models (LLM) have sparked worldwide interest and renewed confidence in the near-future capabilities of Artificial Intelligence (AI), practical feasibility remains an open question.
Indeed, on one side of the spectrum, researchers have argued that the breadth and depth of capabilities exhibited by GPT-4 make it an early version of an AGI \citep{Bubeck2023}.
On the other side, there is extensive evidence that these capabilities may, in fact, not demonstrate any underlying human-level reasoning \citep{Borji2023}.

Furthermore, some argue that the consequences of realizing AGI could lead to drastic societal changes, not all of which would be beneficial to mankind \citep{Ramamoorthy2018}.
This led to entire research fields working to identify and prevent any ``catastrophic risk" that AGI \citep{Sotala2015} could pose.
At the same time, there are growing, opposite, concerns that we humans should not be afraid of AGIs, but instead that \emph{they} should be afraid of \emph{us}.
Authors from both within \citep{Owe2021, Sebo2023} and without \citep{Kateman2024} the scientific community are bringing attention to the need for a moral framework around the use of artificial intelligence.
In the end, though, both sides of the argument may be moot as, according to some authors, ``general artificial intelligence will not be realized" \citep{Fjelland2020} citing critical gaps such as embodiment or social skills \citep{Clocksin2003}.

\begin{figure}
 \begin{subfigure}{.49\columnwidth}
  \includegraphics[width=\textwidth]{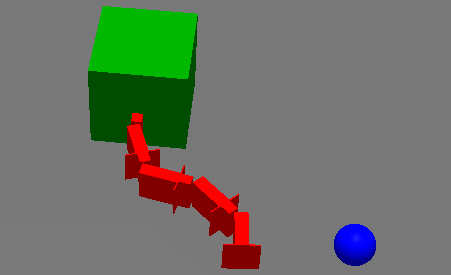}
  \caption{Simulated}
 \end{subfigure}
 \begin{subfigure}{.49\columnwidth}
  \includegraphics[width=\textwidth]{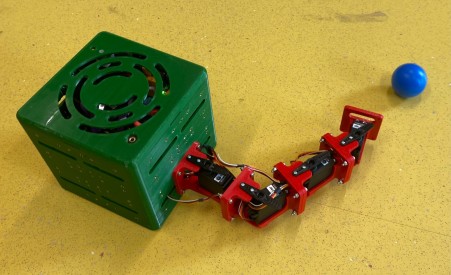}
  \caption{Physical}
 \end{subfigure}
 \caption{
  Thinning the barrier between virtual and real with printable modular robots.
  Prototype video at \url{https://vimeo.com/937579566}.
 }
 \label{vr_barrier}
\end{figure}

\begin{figure*}
 \centering
 \input{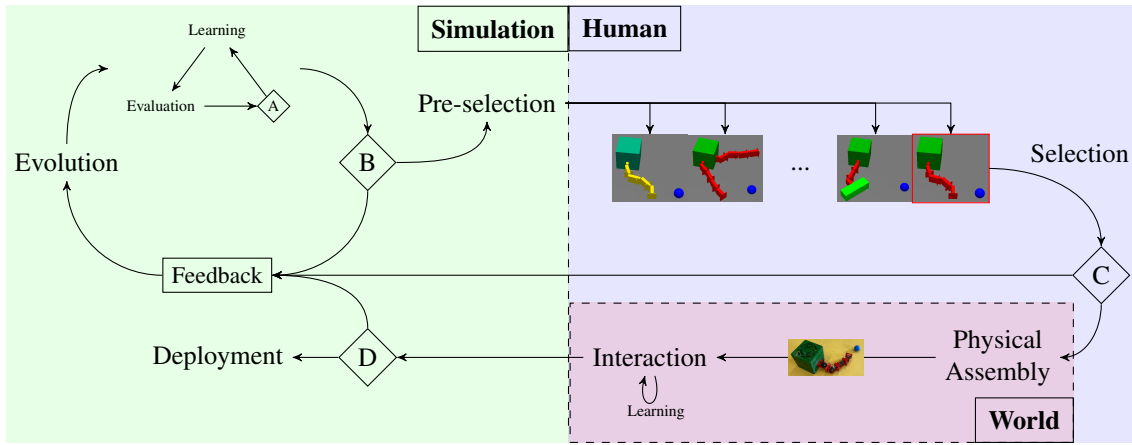}
 \caption{
   Interaction cycles between artificial companions and a human.
   The main evolution loop (left) generates individuals that are evaluated in simulation.
   This process may involve further online optimization, for instance, through (reinforcement) learning.
   Regularly, a pre-selection of promising individuals is presented to a human agent which refines said selection.
   The resulting individuals may then either directly be used as feedback for the evolutionary process or, more infrequently, lead an agent's assembly.
   In the latter case, performance is again measured by a human operator based on multi-modal criteria (autonomy, helpfulness, etc.).
   The interaction is variable in duration and should also include a form of (physical) online optimization to adapt to the reality gap \emph{and} to the human agent.
   Results from this interaction may then be directly used as feedback or, depending on the task and performance, may lead to a functional artificial creature.
   A, B, C, and D correspond to human decision points with increasing levels of controls: from pre-experiment heuristics to extended interactions.
 }
 \label{diagram}
\end{figure*}

One could argue that most contemporary Artificial Intelligence agents are, indeed, disembodied puzzle solvers: \citeauthor{Mnih2013}'s (\citeyear{Mnih2013}) Atari players are just an eye and a hand while Chat-GPT is a browser-based chatbot.
The line of research that we advocate here explicitly considers these limitations (embodiment, mutual trust, and being socially adept) as primary goals, in addition to the agent's ability to self-sustain.
Expert skills, as shown by current AIs, would be an emerging longer-term property, just as they have been in biological evolution.

With this in mind, we suggest that future research in AI should place less emphasis on an agent's puzzle-solving performance, but should instead focus on two intertwined aspects of intelligence:
\begin{description}
 \setlength{\itemsep}{.25em}
 \item [Embodiment] Interacting within a physical, potentially simulated \citep{Kriegman2019}, world ensures a diversity of experience promoting general behavior and adaptivity.
 In fact, we argue that, at its basic level, AI could not be called general without any sensorimotor skills \citep{Pfeifer2006}.
 \item [Social and emotional skills] Most AIs in modern culture are depicted with cold and logical behavior, a position that is reinforced by current implementations.
 In contrast, showing empathy or performing social tasks would not only make artificial agents more relatable, it would also be very likely to improve their own efficiency \citep{Clocksin2003}.
\end{description}

We argue that combining a capacity for (a) autonomous and self-sufficient behavior with (b) human agency, through social directives, would give AGI robust foundations both practically and theoretically.
Furthermore, a very desirable side effect of such an incremental undertaking is that of trust-building, which would start at the individual level.
In essence, the point here is to aim for relatable creatures that humans may, at least \emph{want to}, understand and interact with.

\section{Artificial companions}

In a more practical and short-term view, we propose to take advantage of current works in Evolutionary Robotics to thin the barrier between virtual and real and bring such Artificial General Creatures closer to realization.
Thanks to extensive research in that field, there now exist frameworks in which everything that can be simulated can also be 3D printed \citep{Angus2023, Stuurman2024}, and vice versa.
Reusing inexpensive and readily available components makes it possible to generate complex robot morphologies in virtual space, evaluate performance \emph{a-priori}, and establish them physically with minimal overhead (Figure \ref{vr_barrier}).

To leverage such ``thinning of the barrier", we propose an interactive approach with four nested loops, in an extension of the Triangle of Life \citep{Eiben2013}. Each corresponds to different levels of control, as described in Figure \ref{diagram}.
Starting with a traditional evolutionary loop, individuals are mutated, mated, and evaluated in a virtual environment.
The \emph{a-priori} performance is computed following some additional optimization, for example, through reinforcement learning, until some criteria A is met.
Regularly, a pre-selection (B) of the most promising individuals in the current population is performed based on task-specific criteria (raw performance), but also on the agent's resilience (self-sustainability) and capacity to learn (learning delta, \citet{Miras2020, Luo2022}).
This subset is further evaluated by a ``end-user" human agent, and a refined selection is extracted (C).
The use of an ``end user" is essential to guarantee adequacy between the agents' theoretical performances, as exhibited during simulation, and the expected behavior once deployed in the field.

The result of this second evaluation can then be used in one of two ways: either to provide additional directions to the evolutionary algorithm (e.g., tweaking the fitness, budgets, or population) or to proceed with the assembly of promising agent(s).
Once in the physical world, the agent(s) would interact with humans according to the specifics of the task, while once again, undergoing some form of reinforcement.
Through this, it would adapt to this new environment on two fronts: sensorimotor discrepancies (reality gap) and human behavior (personalizing).
Following this interaction, the results could be directly fed back to the evolutionary algorithm as a final layer of control or, once the performance is deemed adequate, lead to direct deployment (D).

\section{Expected impacts}

On an implementation level, this line of research is expected to foster strong interactions between a host of disciplines ranging from Evolutionary Robotics to Neurosciences.
\begin{description}
 \item [NeuroEvolution and Reinforcement Learning] Due to the emerging morphology of these artificial companions, it is not possible to devise an adequate controller beforehand.
 In contrast, observing the impact of such an interactive multilayered evolution on ANNs' topological features could, in itself, be the subject of dedicated inquiries, e.g. through VfMRI analysis \citep{GodinDubois2023}.
 Additionally, as individuals are expected to learn both in the simulation and in the physical world, they should be endowed with an efficient mechanism to do so, one that is tailored to non-regular topologies.
 As such, further cooperation between NeuroEvolution and Reinforcement Learning researchers could prove most fruitful as advocated in \citep{Stanley2019}.
 
 \item [Neurosciences and Evolutionary Robotics] Similarly, the challenges of embodied Artificial Intelligence bring about the opportunity to investigate the evolution and development of (primitive) artificial life forms on a larger scale.
 For instance, it would be possible to study sensorimotor processing from the elementary mapping of actions to consequences to more complex patterns when facing social or emotional situations both throughout evolution and during an agent's lifetime.
 In fact, this methodology is reminiscent of a domestication process, although at a much faster pace, enabling the exploration of the evolutionary interactions between mind and body in the face of complex environmental constraints.
 
 \item [Ethics, Legal disciplines, Philosophy, etc.] Naturally, the scope also includes social sciences, as AI touches all aspects of society.
 Thus, it would also fall under the scope of this research line to foster debate around the legal status of the artificial creatures that would result from such work.
 Unlike expert systems, the incremental nature of the agents envisioned here would allow a more progressive integration into society with the overarching objective of promoting harmonious, mutually beneficial partnerships \citep{Akata2020, Pianca2022a}.
\end{description}

\section{Acknowledgements}

This research was funded by the Hybrid Intelligence Center, a 10-year programme funded by the Dutch Ministry of Education, Culture and Science through the Netherlands Organization for Scientific Research, \url{https://hybrid-intelligence-centre.nl}, grant number 024.004.022.

\footnotesize
\bibliographystyle{apalike}
\bibliography{main}

\end{document}